\documentclass{article}
\usepackage[utf8]{inputenc}
\usepackage{authblk}
\usepackage{float}
\usepackage{xcolor}
\usepackage{tikz}
\usepackage{soul}
\usepackage{comment}
\usepackage{smartdiagram}
\usepackage{caption}

\usepackage{arev}
\usetikzlibrary{mindmap}
\usepackage{amsmath}
\usepackage{pdflscape}
\usepackage{hyperref}
\usepackage{subcaption}
\usepackage{graphicx}
\usepackage{rotating}
\usepackage{comment}
\usepackage{multirow}
\usepackage{lineno}
\usepackage{array}
\usepackage{lscape}
\usepackage[sorting=none]{biblatex}
\usepackage{setspace}
\DeclareUnicodeCharacter{0301}{\'{o}}
\usepackage{amssymb}
\usepackage[a4paper, margin=1in]{geometry}
\usepackage{times}

\makeatletter
\renewcommand{\maketitle}{\bgroup\setlength{\parindent}{0pt}
\begin{flushleft}
  \textbf{\@title}

  \@author
\end{flushleft}\egroup
}
\makeatother

\providecommand{\keywords}[1]
{
  \textbf{\textit{Keywords---}} #1
}

\font\myfont=cmr12 at 17pt

\title{\myfont{Re-Visiting Explainable AI Evaluation Metrics to Identify The Most Informative Features}}

\date{}
\author[1,2,3,4]{Ahmed M Salih}
\affil[1]{Department of Population Health Sciences, University of Leicester, University Rd, LE1 7RH, Leicester, UK}
\affil[2]{William Harvey Research Institute, NIHR Barts Biomedical Research Centre, Queen Mary University of London, Charterhouse Square, London, EC1M 6BQ, London, UK}
\affil[3]{Barts Heart Centre, St Bartholomew’s Hospital, Barts Health NHS Trust, London, EC1A 7BE, UK}
\affil[4]{PRIME Lab, Scientific Research Center, University of Zakho, Kurdistan Region, Iraq}

\begin{document}
\maketitle
\thispagestyle{empty} 

\noindent

\begin{abstract}
\noindent
Functionality or proxy-based approach is one of the used approaches to evaluate the quality of explainable artificial intelligence methods. It uses statistical methods, definitions and new developed metrics for the evaluation without human intervention. Among them, Selectivity or RemOve And Retrain (ROAR), and Permutation Importance (PI) are the most commonly used metrics to evaluate the quality of explainable artificial intelligence methods to highlight the most significant features in machine learning models. They state that the model performance should experience a sharp reduction if the most informative feature is removed from the model or permuted.  However, the efficiency of both metrics is significantly affected by multicollinearity, number of significant features in the model and the accuracy of the model. This paper shows with empirical examples that both metrics suffer from the aforementioned limitations. Accordingly, we propose expected accuracy interval (EAI), a metric to predict the upper and lower bounds of the the accuracy of the model when ROAR or IP is implemented. The proposed metric found to be very useful especially with collinear features.  

\end{abstract}
\newpage

\section{Introduction}
Explainable artificial intelligence (XAI) emerged as a set of tools, algorithms and methods to help to understand how a machine learning model reaches a specific prediction. In addition, it helps to reveal what are the pixels in an image or features with tabular data that are significantly affect the model decision and are considered as informative. However, XAI methods as other models have their own limitations which necessitate to evaluate their performance appropriately~\cite{vilone2021notions}.\\
Several approaches were proposed to evaluate the quality of XAI that are human-based evaluation, proxy-based evaluation, and literature-based evaluation. Human-based evaluation indicates evaluating XAI by experts in the domain. Prox-based evaluation refers to evaluating XAI performance using some criterion or statistical methods without including human in the loop. Literature-based evaluation usually compares the outcome of XAI with what is already published in the literature to confirm the validity of XAI outcome~\cite{salih2024review}.\\
Proxy-based evaluation approach is more reliable and cheaper compared to the other approaches of evaluation because it is not subjective, faster and can be applied to any domain. Many proxies were proposed to evaluate the outcome of XAI including Selectivity or RemOve And Retrain (ROAR)~\cite{hooker2019benchmark}. ROAR measures the impact on the model performance when the top feature identified by an XAI method is removed from the model. In other words, if XAI identifies feature \textit{A} as the most informative feature in the model, then the model's performance should decline significantly if the model is re-trained after excluding that feature. Although the proposed proxy does make scene in terms of evaluating the outcome of XAI by linking the model performance with the most informative feature, it suffers from not considering the whole picture of multicollinearity and how it affects the model performance even after removing the most significant one.\\
Multicollinearity is one of the big issue when XAI applied to machine learning models~\cite{olaleye2025multilayer}. Many XAI methods including SHAP (SHapley Additive exPlanations)~\cite{lundberg2017unified} considers the features are independent when calculating the contribution of the features toward the prediction. Accordingly, if SHAP identifies feature \textit{A} as the most significant one, then to what extent the model performance will decline if ROAR is implemented. The model might not experience a sharp reduction in the performance after removing the most informative one because there are still collinear significant features in the model. Accordingly, it is very significant to predict the percentage of declining in the model performance to really reveal the impact of that feature in the model outcome. This paper presents a new measure which provide an interval of the expected accuracy when the most significant one is removed from the model or permuted~\cite{salih2024characterizing}.
\section{State of Art}
\begin{figure}[H]
    \centering
    \includegraphics[width=0.8\linewidth]{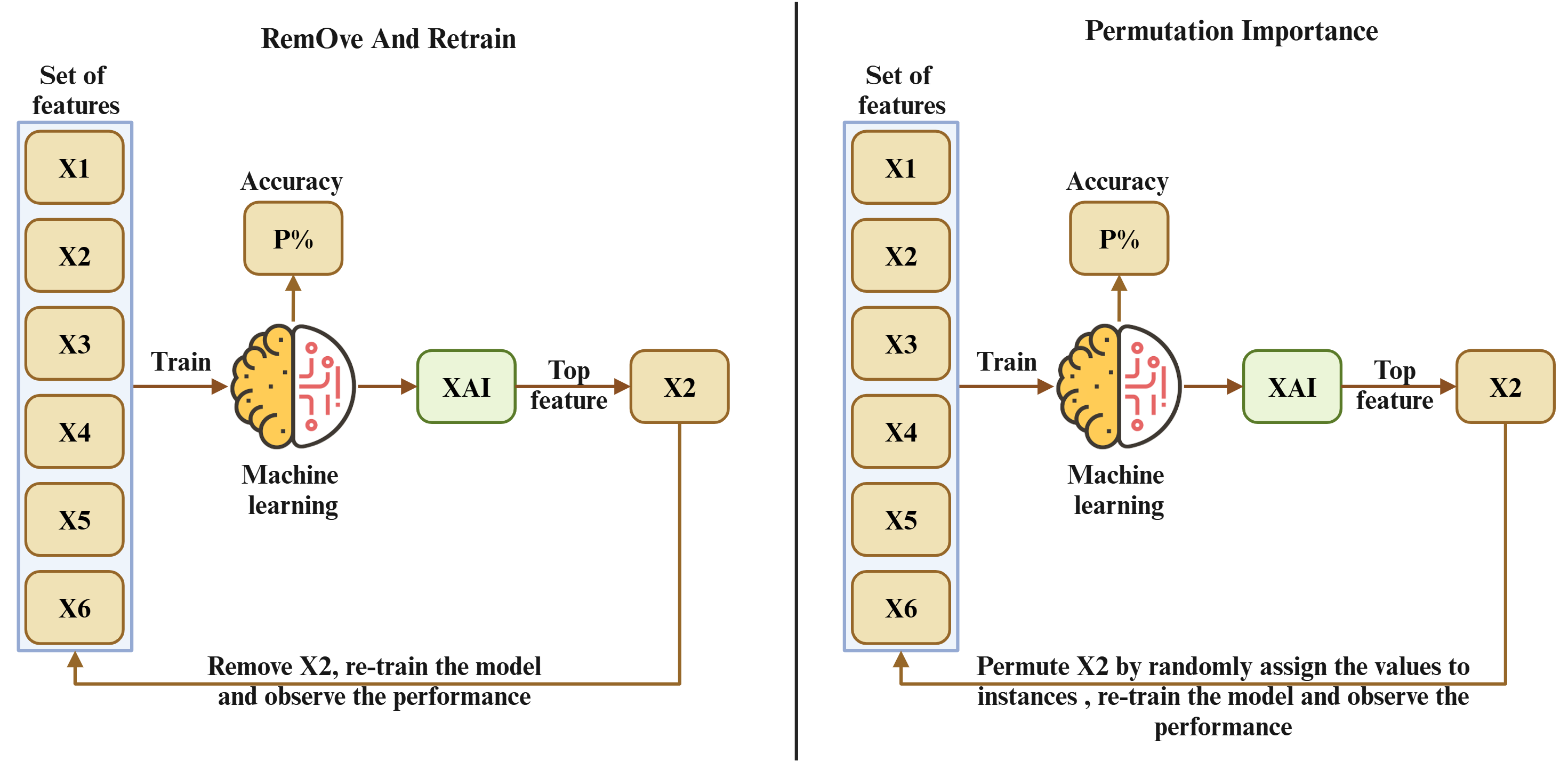}
    \caption{Remove and retain and permutation importance approaches.}
    \label{fi1}
\end{figure}

ROAR~\cite{hooker2019benchmark} and permutation importance (PI)~\cite{altmann2010permutation} are the two most common proxies used to evaluate the performance of any XAI method. Figure~\ref{fi1} shows (on the left) that ROAR indicates the model performance should experience sharp reduction if the top feature identified by any XAI method is removed from the model and re-trained . PI follows similar concept, but instead of removing the top one, it permutes it and expects the model performance will decline (figure~\ref{fi1} on the right). 
\noindent
However, non of them provides a precis numerical measure to what extent the performance of the model might decline after removing/permuting the most informative feature. In addition, the model performance might not decline after removing/permuting the most informative feature if it is correlated with other features that are still in the model and affect the model prediction. Moreover, the model performance might change but not necessary should decline because removing the top feature might help to allow the rest of the features work better in the model due to collinearity. Finally, the two proxies did not provide percentage of the change in the model performance in relation to the model performance before removing the most significant one. It is vital to measure the expected change in the accuracy of the model by considering the current performance of the model and the contribution of each feature toward the model outcome.

\section{Proposed metric}
The proposed method comprises of two main elements which are the accuracy of the current model and the percentage of the contribution of the most significant feature in the model prediction. The accuracy of the current model means the accuracy of the model before removing/permuting the most significant one. In order to calculate the percentage of the contribution of the most significant feature, an XAI should provide a score for each feature. Accordingly, we considered SHAP as an XAI method to explore because it calculates a score for each feature which show how much it contributes in the model prediction.\\
Accordingly, to calculate the percentage of the contribution of the most informative feature:
\begin{equation}
    FCP = SMSF / SSOAF
\end{equation}
where \textit{FCP} is the feature contribution percentage, \textit{SMSF} is the score of the most significant feature and \textit{SSOAF} is the sum of scores of all features in the model. Then, the expected change in the model accuracy can be calculated as:
\begin{equation}
    Expected~\Delta = initial\_acc * FCP
\end{equation}
where~\textit{initial\_acc} is the accuracy of the model before removing/permuting the top feature. Finally, because it is really hard to be precise in predicting the accuracy of any machine learning model, we calculate the upper and lower interval of the expected accuracy after removing/permuting the top feature as:
\begin{equation}
    UI = initial\_acc + Expected~\Delta
\end{equation}
where \textit{UI} is the upper interval of the expected accuracy.
\begin{equation}
    LI = initial\_acc - Expected~\Delta
\end{equation}
where \textit{LI} is the lower interval of the expected accuracy. The interval of the expected accuracy of the model after removing the most significant features is:
\begin{equation}
    EAI = [LI~-~UI]
\end{equation}
where \textit{EAI} is the expected accuracy interval.
\section{Methods and Cases}
\subsection{Real Data}
The dataset of CDC Diabetes Health Indicators~\cite{johnson2014cdc} was downloaded from UCI Archive to perform a binary classification task. The dataset consists of twenty one features while the target was with or without diabetes. The features were mixed of continues and categorical variables for 253,680 samples. We chose 24,000 sample randomly divided equally into with and without diabetes to perform the binary classification. More details about the data can be found~\href{https://archive.ics.uci.edu/dataset/891/cdc+diabetes+health+indicators}{here}.
The Wine Quality dataset from UCI Archive was used to build a linear regression model to predict the quality of wine~\cite{cortez2009modeling}. The dataset consists of ten features for 4,898 instances while the outcome was the quality of the wine starting from 0 to 10. More details about the dataset can be found~\href{https://archive.ics.uci.edu/dataset/186/wine+quality}{here}.
\subsection{Simulated Data}
To develop a binary classification model, \textit{make\_classification} function within the \textit{sklearn$.$datasets} library was used to generate data for 30,000 samples to perform binary classification. The generated data involved twenty features, among them fifteen are informative. The number of samples with class 0 was 14,979 while the number of samples with class 1 was 15,021.
\subsection{Implementation}
The proposed method was implemented in Python language. SHAP was used to explain the model and then extract the SHAP score for each feature. Linear regression model was used to predict the continues values while Logistic regression was used to perform the binary classification. The model was run \textit{n-1} times where \textit{n} is the number of features. For each iteration, the top feature was removed and the model was re-trained and tested again. The whole data was used in the training and test. The regression models were evaluated using coefficient of determination (R2) metric while the classification models were evaluated using F1 score. The data were not normalized neither standardized because the aim is not to improve the model performance, rather to expect the change in the model performance after removing each significant feature. Default parameters of both models were considered.
\section{Results}
\subsection{Real Data}
\begin{figure}[H]
    \centering
    \includegraphics[width=0.9\linewidth]{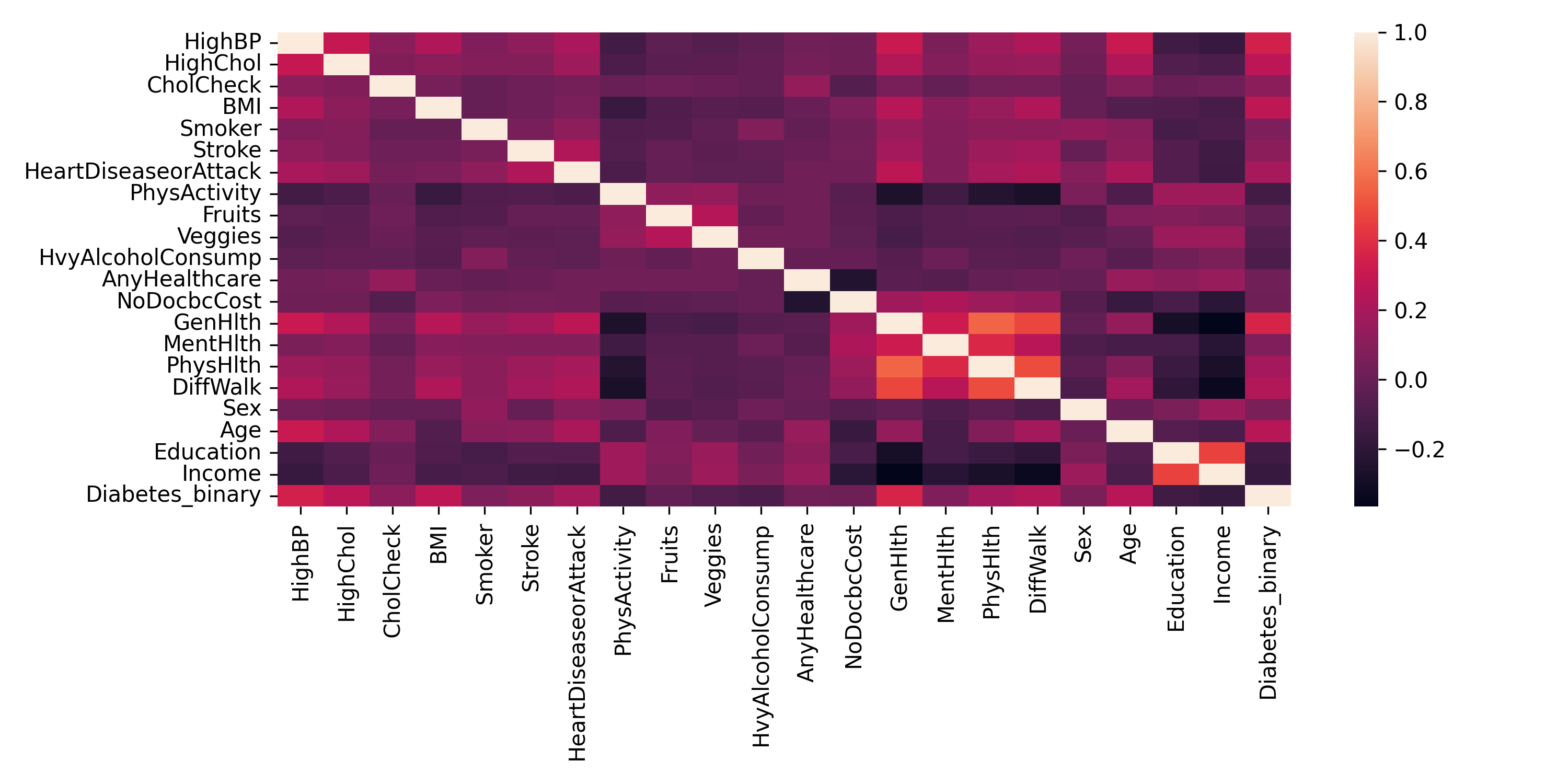}
    \caption{Correlation matrix between the features in the Diabetes dataset.}
    \label{class_real_corr}
\end{figure}

\begin{figure}[H]
    \centering
    \includegraphics[width=0.8\linewidth]{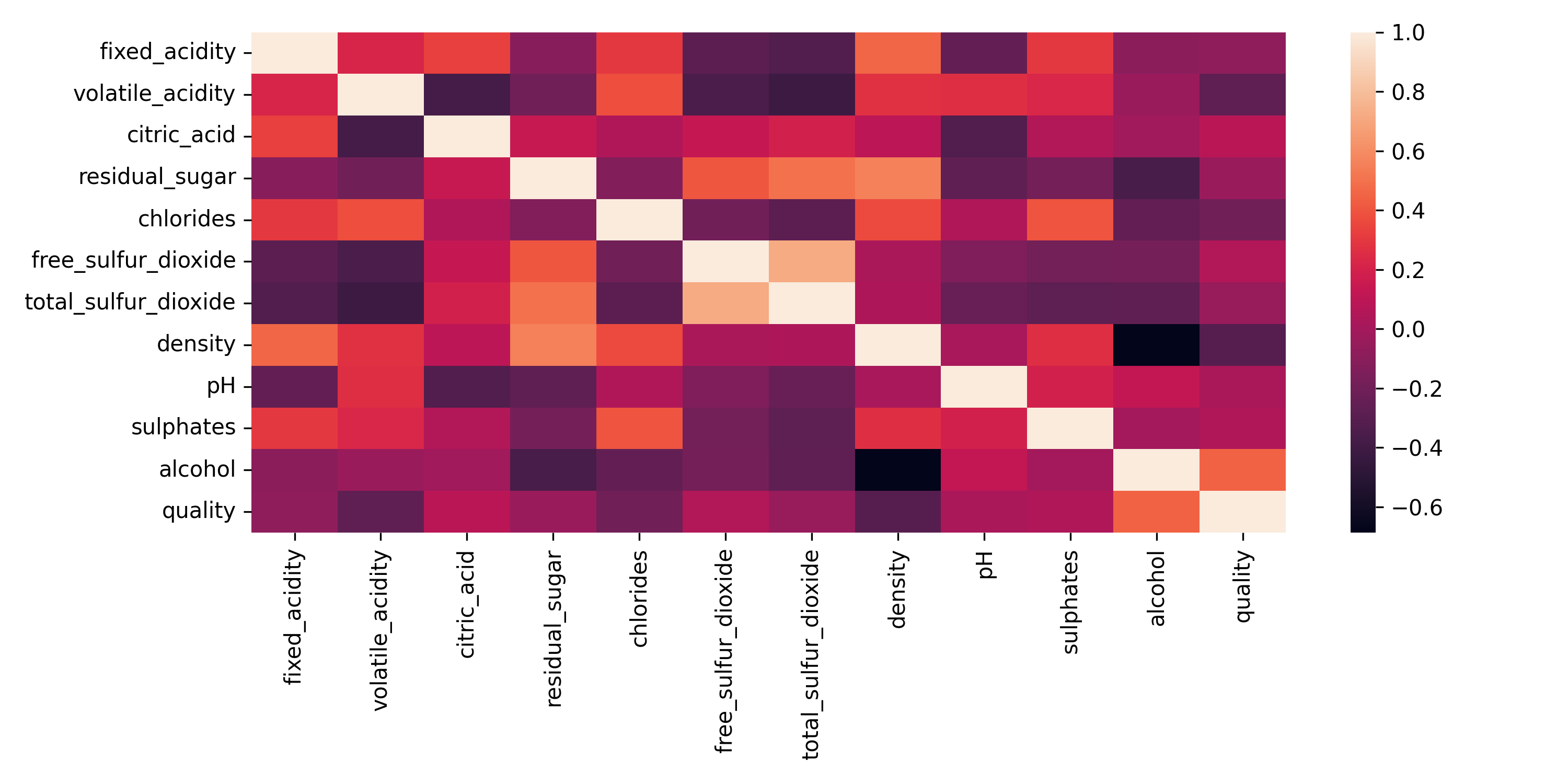}
    \caption{Correlation matrix between the features in the Wine quality dataset.}
    \label{Regre_real_corr}
\end{figure}

Figure~\ref{class_real_corr} shows the correlation between the used features and the outcome. It shows that there are more than one feature including \textit{Income and Education} have significant similar association with the outcome. Accordingly, removing or permuting of the most significant one might not result in a sharp reduction in the model accuracy.
\noindent
Table~\ref{Class_real} lists the most significant features in the model for each iteration, the accuracy of the model and the interval of the expected accuracy after removing the most significant feature. The figure shows the most significant feature based on the SHAP score in the first iteration is~\textit{GenHlth} and the accuracy before removing it is 0.7318. It also shows that the expected accuracy after removing \textit{GenHlth} will fall between 0.5945 and 0.8691. Although after removing the most significant feature based on SHAP score, the changes in the accuracy of the model in the iteration one and two is very small (0.0097). Moreover, the accuracy of the model might increase instead of declining after removing the most informative one as it is shown in seventeen iteration after removing the most significant feature in iteration sixteen which was \textit{Stroke}. The accuracy increased significantly from 0.5164 to 0.6716. Accordingly, ROAR/PI might not work properly in all cases especially when the model performance is already low.
\begin{table}[H]
\small
\centering
\begin{tabular}{|c|l|c|cc|}
\hline
\multicolumn{1}{|l|}{\textbf{Iteration}} & \textbf{MSF}         & \multicolumn{1}{l|}{\textbf{\begin{tabular}[c]{@{}l@{}}Accuracy of\\  the model (F1)\end{tabular}}} & \multicolumn{2}{c|}{\textbf{\begin{tabular}[c]{@{}c@{}}Expected accuracy \\ of  next model\end{tabular}}} \\ \hline
1                                        & GenHlth              & 0.7318                                                                                         & \multicolumn{1}{c|}{0.5945}                                    & 0.8691                                   \\ \hline
2                                        & BMI                  & 0.7221                                                                                         & \multicolumn{1}{c|}{0.5859}                                    & 0.8584                                   \\ \hline
3                                        & HighBP               & 0.7084                                                                                         & \multicolumn{1}{c|}{0.5394}                                    & 0.8775                                   \\ \hline
4                                        & HighChol             & 0.6785                                                                                         & \multicolumn{1}{c|}{0.5269}                                    & 0.8302                                   \\ \hline
5                                        & Age                  & 0.654                                                                                          & \multicolumn{1}{c|}{0.5042}                                    & 0.8038                                   \\ \hline
6                                        & DiffWalk             & 0.6191                                                                                         & \multicolumn{1}{c|}{0.5083}                                    & 0.73                                     \\ \hline
7                                        & HeartDiseaseorAttack & 0.6117                                                                                         & \multicolumn{1}{c|}{0.5163}                                    & 0.7071                                   \\ \hline
8                                        & PhysHlth             & 0.6069                                                                                         & \multicolumn{1}{c|}{0.4979}                                    & 0.716                                    \\ \hline
9                                        & Income               & 0.6085                                                                                         & \multicolumn{1}{c|}{0.487}                                     & 0.73                                     \\ \hline
10                                       & PhysActivity         & 0.5953                                                                                         & \multicolumn{1}{c|}{0.4863}                                    & 0.7043                                   \\ \hline
11                                       & Education            & 0.5818                                                                                         & \multicolumn{1}{c|}{0.4467}                                    & 0.7169                                   \\ \hline
12                                       & Sex                  & 0.5552                                                                                         & \multicolumn{1}{c|}{0.4523}                                    & 0.6581                                   \\ \hline
13                                       & Smoker               & 0.6076                                                                                         & \multicolumn{1}{c|}{0.4632}                                    & 0.752                                    \\ \hline
14                                       & MentHlth             & 0.493                                                                                          & \multicolumn{1}{c|}{0.3881}                                    & 0.5978                                   \\ \hline
15                                       & Veggies              & 0.4457                                                                                         & \multicolumn{1}{c|}{0.3353}                                    & 0.5562                                   \\ \hline
16                                       & Stroke               & 0.5164                                                                                         & \multicolumn{1}{c|}{0.3931}                                    & 0.6397                                   \\ \hline
17                                       & HvyAlcoholConsump    & \textcolor{red}{0.6716}                                                                                         & \multicolumn{1}{c|}{0.4657}                                    & 0.8774                                   \\ \hline
18                                       & CholCheck            & 0.4843                                                                                         & \multicolumn{1}{c|}{0.3072}                                    & 0.6615                                   \\ \hline
19                                       & Fruits               & 0.4772                                                                                         & \multicolumn{1}{c|}{0.2952}                                    & 0.6591                                   \\ \hline
20                                       & AnyHealthcare        & \textcolor{red}{0.1513}                                                                                         & \multicolumn{1}{c|}{}                                    &                                   \\ \hline
\end{tabular}
\caption{Models performance when real data was used to perform binary classification. Those highlighted in red are outside of the interval of the expected accuracy. MSF: most significant feature.}
\label{Class_real}
\end{table}
\noindent
Figure~\ref{Regre_real_corr} shows the correlation matrix between the features of the real data to predict the quality of wine. It shows there are more than one feature have similar association with the outcome. For instance, \textit{sulphates, total\_sulfur\_dioxide and citric\_acid} have similar association which they might affect the model performance similarly.
\noindent
Table~\ref{Reg_real} shows the most informative features, the accuracy of the model and the expected interval of the accuracy for each iteration. It shows there was not a sharp reduction when the most significant features (\textit{alcohol}) was removed from the model because the second feature (\textit{density}) seems also significantly is associated with the outcome. The changes in the model performance after removing the top one got smaller after iteration five and even arbitrary because the model performance is not good enough to apply any XAI method to reveal the contributions of the features toward the outcome.
\begin{table}[H]
\small
\centering
\begin{tabular}{|c|l|c|cc|}
\hline
\multicolumn{1}{|l|}{\textbf{Iteration}} & \textbf{MSF}           & \multicolumn{1}{l|}{\textbf{\begin{tabular}[c]{@{}l@{}}Accuracy of\\  the model (R2)\end{tabular}}} & \multicolumn{2}{c|}{\textbf{\begin{tabular}[c]{@{}c@{}}Expected accuracy \\ of  next model\end{tabular}}} \\ \hline
1                                        & alcohol                & 0.2921                                                                                         & \multicolumn{1}{c|}{0.2243}                                    & 0.36                                     \\ \hline
2                                        & density                & 0.2643                                                                                         & \multicolumn{1}{c|}{0.1831}                                    & 0.3456                                   \\ \hline
3                                        & total\_sulfur\_dioxide & 0.1453                                                                                         & \multicolumn{1}{c|}{0.105}                                     & 0.1855                                   \\ \hline
4                                        & volatile\_acidity      & 0.1128                                                                                         & \multicolumn{1}{c|}{0.0733}                                    & 0.1524                                   \\ \hline
5                                        & chlorides              & 0.0757                                                                                         & \multicolumn{1}{c|}{0.0536}                                    & 0.0978                                   \\ \hline
6                                        & citric\_acid           & \textcolor{red}{0.0293}                                                                                         & \multicolumn{1}{c|}{0.0208}                                    & 0.0377                                   \\ \hline
7                                        & fixed\_acidity         & \textcolor{red}{0.0156}                                                                                         & \multicolumn{1}{c|}{0.0114}                                    & 0.0198                                   \\ \hline
8                                        & free\_sulfur\_dioxide  & \textcolor{red}{0.0091}                                                                                         & \multicolumn{1}{c|}{0.005}                                     & 0.0132                                   \\ \hline
9                                        & residual\_sugar        & \textcolor{red}{0.0024}                                                                                         & \multicolumn{1}{c|}{0.0013}                                    & 0.0035                                   \\ \hline
10                                       & sulphates              & 0.0016                                                                                        & \multicolumn{1}{c|}{}                                    &                                  \\ \hline
\end{tabular}
\caption{Models performance when real data was used to perform the regression. Those highlighted in red are outside of the interval of the expected accuracy. MSF: most significant feature.}
\label{Reg_real}
\end{table}

\subsection{Simulated Data}
\begin{figure}[H]
    \centering
    \includegraphics[width=0.85\linewidth]{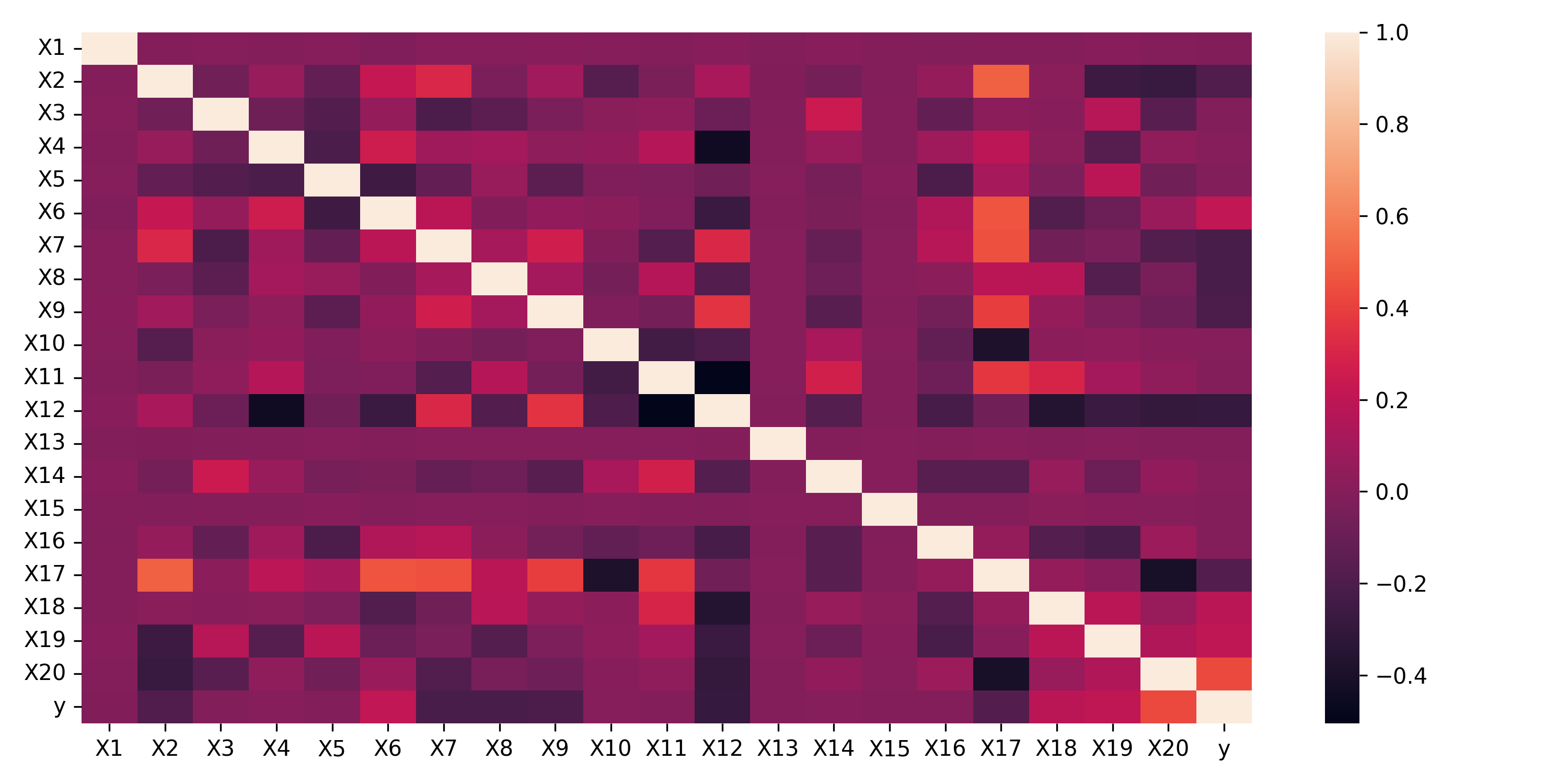}
    \caption{Correlation matrix between the features in the simulated dataset to perform binary classification.}
    \label{Class_simul_corr}
\end{figure}

Similar pattern is observed with the simulated data. Many features are collinear and at the same time are associated with the outcome as it is shown in figure~\ref{Class_simul_corr}. The figure shows the correlation matrix between the simulated features to perform a binary classification task. It shows there is multicollinearity among the features and with the outcome which might affect how ROAR/PI work.
\noindent
Table~\ref{Class_sim} shows the model performance in each iteration. It shows that the model performance did not change in the first three iterations although the most informative feature is removed in each iteration. However, the interval of the expected accuracy is changed in each iteration because the contribution of the most significant feature is changed in each iteration. The model accuracy keeps declining till iteration fourteen and then increases gradually again till it reached an accuracy with one feature in the model similar to iteration ten.
\begin{table}[H]
\small
\centering
\begin{tabular}{|c|c|c|cc|}
\hline
\multicolumn{1}{|l|}{\textbf{Iteration}} & \textbf{MSF}             & \multicolumn{1}{l|}{\textbf{\begin{tabular}[c]{@{}l@{}}Accuracy of\\ the model (F1)\end{tabular}}} & \multicolumn{2}{l|}{\textbf{\begin{tabular}[c]{@{}l@{}}Expected accuracy\\ of next model\end{tabular}}} \\ \hline
1                                        & \multicolumn{1}{c|}{X6}  & 0.7899                                                                                             & \multicolumn{1}{c|}{0.6754}                                   & 0.9044                                  \\ \hline
2                                        & \multicolumn{1}{c|}{X17} & 0.7899                                                                                             & \multicolumn{1}{c|}{0.6926}                                   & 0.8871                                  \\ \hline
3                                        & \multicolumn{1}{c|}{X12} & 0.7899                                                                                             & \multicolumn{1}{c|}{0.6026}                                   & 0.9773                                  \\ \hline
4                                        & \multicolumn{1}{c|}{X20} & 0.7539                                                                                             & \multicolumn{1}{c|}{0.5574}                                   & 0.9504                                  \\ \hline
5                                        & \multicolumn{1}{c|}{X18} & 0.7065                                                                                             & \multicolumn{1}{c|}{0.5982}                                   & 0.8147                                  \\ \hline
6                                        & \multicolumn{1}{c|}{X19} & 0.6812                                                                                             & \multicolumn{1}{c|}{0.5689}                                   & 0.7936                                  \\ \hline
7                                        & \multicolumn{1}{c|}{X8}  & 0.6681                                                                                             & \multicolumn{1}{c|}{0.524}                                    & 0.8122                                  \\ \hline
8                                        & \multicolumn{1}{c|}{X9}  & 0.6419                                                                                             & \multicolumn{1}{c|}{0.5015}                                   & 0.7822                                  \\ \hline
9                                        & \multicolumn{1}{c|}{X7}  & 0.6158                                                                                             & \multicolumn{1}{c|}{0.4184}                                   & 0.8131                                  \\ \hline
10                                       & \multicolumn{1}{c|}{X2}  & 0.5779                                                                                             & \multicolumn{1}{c|}{0.2725}                                   & 0.8833                                  \\ \hline
11                                       & X1                       & 0.5212                                                                                             & \multicolumn{1}{c|}{0.3942}                                   & 0.6482                                  \\ \hline
12                                       & X14                      & 0.5227                                                                                             & \multicolumn{1}{c|}{0.4247}                                   & 0.6206                                  \\ \hline
13                                       & X5                       & 0.5177                                                                                             & \multicolumn{1}{c|}{0.4108}                                   & 0.6247                                  \\ \hline
14                                       & X4                       & 0.5171                                                                                             & \multicolumn{1}{c|}{0.3912}                                   & 0.643                                   \\ \hline
15                                       & X3                       & 0.5277                                                                                             & \multicolumn{1}{c|}{0.3872}                                   & 0.6682                                  \\ \hline
16                                       & X10                      & 0.5288                                                                                             & \multicolumn{1}{c|}{0.3771}                                   & 0.6806                                  \\ \hline
17                                       & X11                      & 0.5391                                                                                             & \multicolumn{1}{c|}{0.3622}                                   & 0.7159                                  \\ \hline
18                                       & X15                      & 0.5572                                                                                             & \multicolumn{1}{c|}{0.3234}                                   & 0.7911                                  \\ \hline
19                                       & X13                      & 0.5741                                                                                             & \multicolumn{1}{c|}{}                                    &                                 \\ \hline
\end{tabular}
\caption{Models performance when simulated data was used to perform the binary classification task. MSF: most significant feature.}
\label{Class_sim}
\end{table}
\section{Discussion}
ROAR and PI importance are the most common used XAI proxies to assess the quality of the machine learning models to correctly identify the most informative features in tabular data or the pixels in an image~\cite{mundhenk2019efficient}~\cite{shevskaya2021explainable}. The main concept of the proxies matches with the aims of XAI. However, our results show that both proxies suffer from some limitations which necessitate a careful consideration when they are adopted to evaluate the quality of any XAI method. If there are more than one significant feature in the model, then even removing some of them might not lead to decline in the model performance. In addition, the results show that in some cases the model performance might increase instead of decreasing because the left features might work better and improve the accuracy of the model when the top one is absent. Moreover, if the performance of the model is already low, then removing the most informative features might not affect the performance of the model or the change might be tiny.\\
In this work, we proposed a simple yet useful and informative metric to predict the interval of the predicted accuracy. A wider interval might be interpreted as after removing the most significant features from the model, the new significant one has greater impact on the predicted outcome. On the contrary, the narrower the interval indicates that the new significant feature has a similar impact to the one removed on the outcome.\\
Such interval indeed is useful in some applications where there is a high degree of collinearity between the features. For instance, diabetes, smoking, alcohol and hypertension are vascular risk factors which increases the chance of cardiovascular diseases~\cite{ciumuarnean2021cardiovascular}. When the predicted interval in such case is wider after removing the most informative feature, this indicates that the new significant (e.g., smoking) feature has greater impact and risk of cardiovascular diseases. This is specifically useful with any XAI method that considers the features are independent when it explains any machine learning model.\\
The proposed metric might has some limitations which are more related to the used data and the applied XAI method. For instance, it works only for those XAI methods that provide a score for each feature which represent the contribution of each feature toward the outcome.I addition, if the models work perfectly either because of overfitting or because the features are capable perfectly (100\%) to predict the outcome, then the prediction interval of the expected accuracy of the first and the second iterations will not be precise because the accuracy will be 100\% multiplied with the contribution of the feature. 
\section{Acknowledgments}
AMS acknowledges support from The Leicester City Football Club (LCFC). Figure 1 is generated by Biorender~(\url{https://www.biorender.com}).
\section{Data availability}
The real datasets used are available at \href{https://archive.ics.uci.edu/}{UCI Machine Learning Repository}. The simulated dataset can be downloaded from the supplementary.

\printbibliography

\end{document}